\documentclass{bmcart}

\usepackage{amsthm,amsmath}
\usepackage{url}
\RequirePackage{hyperref}
\usepackage[utf8]{inputenc} %unicode support

\usepackage{algorithm}
\usepackage[noend]{algpseudocode}
\usepackage{hyphenat}

\usepackage{graphicx}

\startlocaldefs
\endlocaldefs

\begin{document}

%%% Start of article front matter
\begin{frontmatter}

\begin{fmbox}
\dochead{Research}


\title{Combined Pruning for Nested Cross-Validation to Accelerate Automated Hyperparameter Optimization for Embedded Feature Selection in High-Dimensional Data with Very Small Sample Sizes}

%%%%%%%%%%%%%%%%%%%%%%%%%%%%%%%%%%%%%%%%%%%%%%
%%                                          %%
%% Enter the authors here                   %%
%%                                          %%
%% Specify information, if available,       %%
%% in the form:                             %%
%%   <key>={<id1>,<id2>}                    %%
%%   <key>=                                 %%
%% Comment or delete the keys which are     %%
%% not used. Repeat \author command as much %%
%% as required.                             %%
%%                                          %%
%%%%%%%%%%%%%%%%%%%%%%%%%%%%%%%%%%%%%%%%%%%%%%

\author[
  addressref={aff1,aff2,aff3},                   % id's of addresses, e.g. {aff1,aff2}
% noteref={n1},                        % id's of article notes, if any
  email={s.may@ostfalia.de}   % email address
]{\inits{S.M.}\fnm{Sigrun} \snm{May}}
\author[
  addressref={aff2},                   % id's of addresses, e.g. {aff1,aff2}
  email={sven.hartmann@tu-clausthal.de}   % email address
]{\inits{S.H.}\fnm{Sven} \snm{Hartmann}}
\author[
  addressref={aff1,aff3},
  corref={aff1},                       % id of corresponding address, if any
  email={f.klawonn@ostfalia.de}
]{\inits{F.K.}\fnm{Frank} \snm{Klawonn}}

%%%%%%%%%%%%%%%%%%%%%%%%%%%%%%%%%%%%%%%%%%%%%%
%%                                          %%
%% Enter the authors' addresses here        %%
%%                                          %%
%% Repeat \address commands as much as      %%
%% required.                                %%
%%                                          %%
%%%%%%%%%%%%%%%%%%%%%%%%%%%%%%%%%%%%%%%%%%%%%%

\address[id=aff1]{%                           % unique id
  \orgdiv{Department of Computer Science},             % department, if any
  \orgname{Ostfalia University of Applied Sciences},          % university, etc
  \city{Wolfenbüttel},                              % city
  \cny{Germany}                                    % country
}
\address[id=aff2]{%
  \orgdiv{Department of Informatics},
  \orgname{Clausthal University of Technology},
  %\street{},
  %\postcode{}
  \city{Clausthal-Zellerfeld},
  \cny{Germany}
}
\address[id=aff3]{%
  \orgdiv{Biostatistics Research Group},
  \orgname{Helmholtz Centre for Infection Research},
  %\street{},
  %\postcode{}
  \city{Braunschweig},
  \cny{Germany}
}

%%%%%%%%%%%%%%%%%%%%%%%%%%%%%%%%%%%%%%%%%%%%%%
%%                                          %%
%% Enter short notes here                   %%
%%                                          %%
%% Short notes will be after addresses      %%
%% on first page.                           %%
%%                                          %%
%%%%%%%%%%%%%%%%%%%%%%%%%%%%%%%%%%%%%%%%%%%%%%

%%\note{Sample of title note}     % note to the article
%\note[id=n1]{Equal contributor} % note, connected to author

\end{fmbox}% comment this for two column layout

%%%%%%%%%%%%%%%%%%%%%%%%%%%%%%%%%%%%%%%%%%%%%%
%%                                          %%
%% The Abstract begins here                 %%
%%                                          %%
%% Please refer to the Instructions for     %%
%% authors on http://www.biomedcentral.com  %%
%% and include the section headings         %%
%% accordingly for your article type.       %%
%%                                          %%
%%%%%%%%%%%%%%%%%%%%%%%%%%%%%%%%%%%%%%%%%%%%%%

\begin{abstractbox}

\begin{abstract} % abstract
\parttitle{Background}
Embedded feature selection in high-dimensional data with very small sample sizes requires optimized hyperparameters for the model building process. For this hyperparameter optimization, nested cross-validation must be applied to avoid a biased performance estimation. The resulting repeated training with high-dimensional data leads to very long computation times. Moreover, it is likely to observe a high variance in the individual performance evaluation metrics caused by outliers in tiny validation sets. Therefore, early stopping applying standard pruning algorithms to save time risks discarding promising hyperparameter sets.
\parttitle{Result}
To speed up feature selection for high-dimensional data with tiny sample size, we adapt the use of a state-of-the-art asynchronous successive halving pruner. In addition, we combine it with two complementary pruning strategies based on domain or prior knowledge. One pruning strategy immediately stops computing trials with semantically meaningless results for the selected hyperparameter combinations. 
The other is a new extrapolating threshold pruning strategy suitable for nested-cross-validation with a high variance of performance evaluation metrics. In repeated experiments, our combined pruning strategy keeps all promising trials. At the same time, the calculation time is substantially reduced compared to using a state-of-the-art asynchronous successive halving pruner alone. Up to 81.3\% fewer models were trained achieving the same optimization result.
\parttitle{Conclusion}
The proposed combined pruning strategy accelerates data analysis or enables deeper searches for hyperparameters within the same computation time. This leads to significant savings in time, money and energy consumption, opening the door to advanced, time-consuming analyses.
\end{abstract}

%%%%%%%%%%%%%%%%%%%%%%%%%%%%%%%%%%%%%%%%%%%%%%
%%                                          %%
%% The keywords begin here                  %%
%%                                          %%
%% Put each keyword in separate \kwd{}.     %%
%%                                          %%
%%%%%%%%%%%%%%%%%%%%%%%%%%%%%%%%%%%%%%%%%%%%%%


% MSC classifications codes, if any
%\begin{keyword}[class=AMS]
%\kwd[Primary ]{}
%\kwd{}
%\kwd[; secondary ]{}
%\end{keyword}

\end{abstractbox}
%
%\end{fmbox}% uncomment this for twcolumn layout

\end{frontmatter}

%%%%%%%%%%%%%%%%%%%%%%%%%%%%%%%%%%%%%%%%%%%%%%
%%                                          %%
%% The Main Body begins here                %%
%%                                          %%
%% Please refer to the instructions for     %%
%% authors on:                              %%
%% http://www.biomedcentral.com/info/authors%%
%% and include the section headings         %%
%% accordingly for your article type.       %%
%%                                          %%
%% See the Results and Discussion section   %%
%% for details on how to create sub-sections%%
%%                                          %%
%% use \cite{...} to cite references        %%
%%  \cite{koon} and                         %%
%%  \cite{oreg,khar,zvai,xjon,schn,pond}    %%
%%  \nocite{smith,marg,hunn,advi,koha,mouse}%%
%%                                          %%
%%%%%%%%%%%%%%%%%%%%%%%%%%%%%%%%%%%%%%%%%%%%%%

%%%%%%%%%%%%%%%%
%% Background %%
%%
\section*{Background}
Embedded feature selection in high-dimensional data to reduce the feature dimension can be time-consuming or even unmanageable in a reasonable time without acceleration. Such high-dimensional data with a very small sample size occur, for example, in biomarker pilot studies\footnote{Biomarkers are measurable indicators (1) for medical risk factors, (2) for a biological condition, (3) to study a disease, (4) to predict a diagnosis, (5) to determine the state of a disease or (6) the effectiveness of a treatment \cite{Califf2018}. Because of their usefulness and broad applicability, the search for biomarkers has gained considerable interest. Small pilot studies can be a first step when searching for new biomarkers or biomarker combinations to save effort, money, and time. 
If data within a biomarker pilot study is based on high-throughput technologies, sample generation can be both expensive and time-consuming. Therefore, the sample size in these pilot studies is usually very limited while the number of potential biomarker candidates is easily in the thousands. \cite{Klawonn2016}}. For this kind of research, feature selection aims to exclude irrelevant features. When analyzing high-dimensional data with a statistically insufficient sample size, random effects cannot be eliminated \cite{Al-Mekhlafi2022}, overfitting is difficult to avoid, and outliers can have a high impact. Therefore, it is unlikely to directly find a final biomarker (feature) combination for clinical use within a pilot study. However, substantial feature reduction is highly relevant for extended studies with a larger sample size \cite{Al-Mekhlafi2022}. For example, only a massively reduced feature dimension allows the application of targeted "omics" \cite{Coman2020}, such as targeted proteomics \cite{Marx2013}, to detect proteins of interest with high sensitivity, quantitative accuracy, and reproducibility.\\ 
One approach to achieve a reduction of the feature dimension is embedded feature selection \cite{Ma2008} \cite{Torres2019} \cite{Qi2012}. How well the model fits the data is measured with a performance evaluation metric. For small sample sizes this is usually determined using cross-validation. But if only a very limited number of samples is available, k-fold cross-validation produces strongly biased performance estimates  \cite{Vabalas2019}, \cite{Cawley2010}. Vabalas et. al \cite{Vabalas2019} therefore suggest applying nested cross-validation instead to obtain robust and unbiased performance estimates regardless of the sample size. The nested cross-validation has an outer and an inner loop. Thus, a complete nested cross validation has the number of outer folds multiplied by the number of inner folds (\textit{folds}$_{outer-cv} * {\textit{folds}}_{inner-cv}$) steps. In each of these steps a new model is built. The runtime of the model building process increases with the number of features. Consequently, building a large number of models with a high feature dimension leads to a long overall computation time.\\
To better adapt a machine learning model to its specific task, its hyperparameters must be tuned. Choosing tuned hyperparameters can directly impact the performance of the model \cite{Yang2020}. In order to reduce overfitting it is advisable to regularize the model building process by adapting the corresponding hyperparameters \cite{Nordhausen2009}. Given the multitude of hyperparameter dimensions that can be optimized, the search space for hyperparameter combinations can be large, making it challenging to evaluate every optimization configuration. Hence, finding a good combination of hyperparameter values often requires many trials. For each trial, nested cross-validation has to be applied. In addition, further time-consuming analyses may be necessary, such as computing SHAP values \cite{Lundberg2020} to determine feature importances \cite{Marcilio2020} based on the trained models. Hence, long computation times for each trial are a bottleneck for the required hyperparameter optimization.\\
One solution to this problem is to stop trials with unfavourable hyperparameter values at an early stage. 
The strategy of early stopping in hyperparameter optimization - also called pruning - is different from the early stopping used by machine learning algorithms against overfitting. In this context, it means identifying and terminating unpromising trials of hyperparameter values as early as possible while continuing to calculate with the most promising combinations. By stopping the least promising trials early during training, better feature selection can be achieved with the same computational power. With the ability to compute more trials in less time, the hyperparameter space can be explored faster or deeper. This acceleration saves time and energy - and thus also money. Therefore, pruning is essential when budget is limited and calculation times are long.\\
However, standard pruning algorithms cannot handle the considered data in the usual way. They are unreliable at an early stage of nested cross-validation due to the high variance of the performance evaluation metrics (see Figure \ref{cumulated_mean}).
To solve this problem, we propose a new combined pruning strategy to accelerate this specific hyperparameter optimization massively. To the best of our current knowledge, no specific or adapted pruner currently exists suitable for nested cross-validation with high variance in the performance metric. Our proposed pruning strategy is the first to consider the treatment of outliers in small validation sets and the semantics of the results. In addition, we incorporate prior knowledge and combine different pruning approaches. Wang et. al \cite{Wang2021} also consider prior knowledge for pruning. They gain procedural knowledge from previous configurations and determine important hyperparameters to derive optimal configurations for neural networks. In contrast to Wang et. al \cite{Wang2021}, our approach refers to the minimum requirements for the performance evaluation metric. Moreover, we adjust the use of standard pruning strategies. Finally, we combine the three different approaches.\\

\section*{Pruning Automated Hyperparameter Optimization} \label{pruner}
We propose a three-layer pruning algorithm (see Algorithm \ref{alg: three-layer}) that accelerates the hyperparameter optimization.
\begin{algorithm}
\begin{algorithmic}[1]
\For{\textit{every iteration of the outer cross-validation}} \Comment{nested cross-validation}
\For{\textit{every iteration of the inner cross-validation}}
\State build model
\State evaluate model on validation set
\\
\Comment{pruning}
\If{\textit{no feature is important}}
\State stop trial for current combination of hyperparemter values
\State (see \textit{\nameref{semantic pruning}})
\\
\ElsIf {\textit{extrapolated performance evaluation metric\\ \hspace*{23mm} is not better than a user defined threshold }}
\State stop trial for current combination of hyperparemter values
\State (see \textit{\nameref{threshold-pruner}})
\\
\ElsIf {\textit{comparison-based standard pruner would prune }}
\State stop trial for current combination of hyperparemter values
\State (see \textit{\nameref{standard pruning strategy}})
\\
\Else
\State continue
\EndIf
\\
\State evaluate model on test set of outer loop
\EndFor \label{inner feature selection loop}
\EndFor \label{test loop}
\caption{Combined three-layer pruning.} 
\label{alg: three-layer}
\end{algorithmic}
\end{algorithm}
\begin{samepage}
The three combined parts prune trials if they 
\begin{itemize}
\item are worse compared to previously calculated trials\\
(see \textit{\nameref{standard pruning strategy}}),
\item are semantically irrelevant due to lack of information for feature selection\\
(see \textit{\nameref{semantic pruning}}), or
\item have insufficient prediction quality for the specific use case\\
(see \textit{\nameref{threshold-pruner}}).
\end{itemize}
\end{samepage}
In the following subsections, we present the three different concepts in detail. We then describe the combined three-layer pruner.
\subsection*{Adapting Intermediate Values for Standard Pruning Algorithms} \label{standard pruning strategy}
\begin{figure}
\includegraphics[width=122mm]{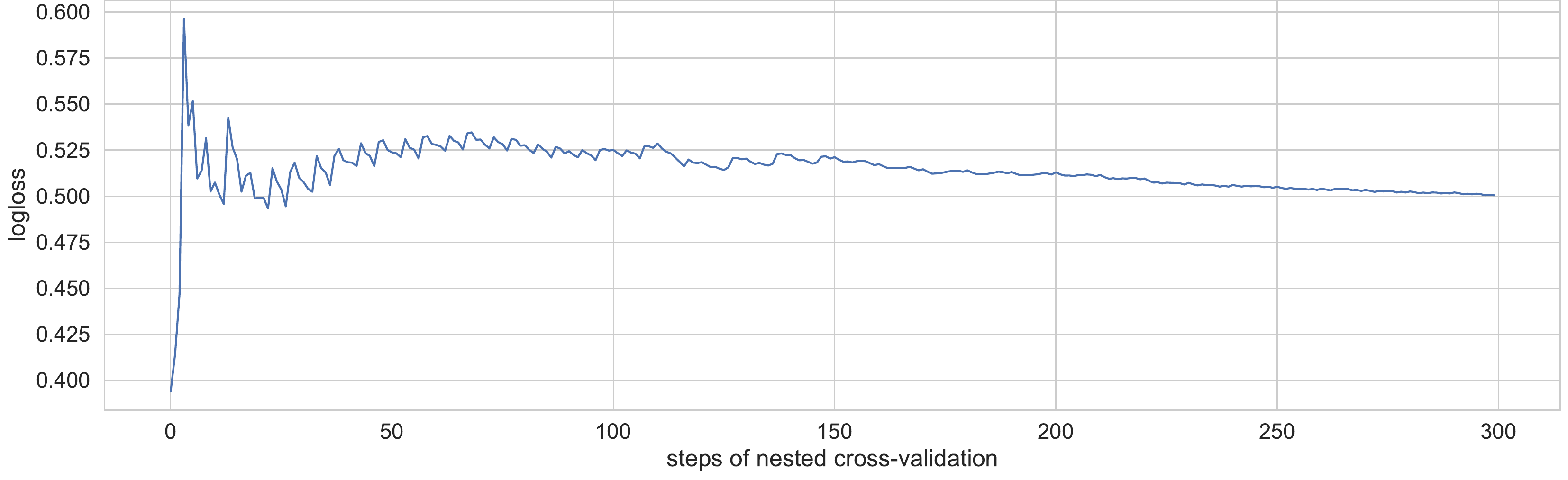}
\caption{State-of-the-art cross-validation: Average of each previously calculated performance evaluation metric of a nested cross-validation as intermediate values. The figure shows a single trial calculated with colon cancer data \cite{Alon1999} (see \nameref{experimental-setup}. Experimental Setup).} 
\label{cumulated_mean}
\end{figure}

\begin{figure}
\includegraphics[width=122mm]{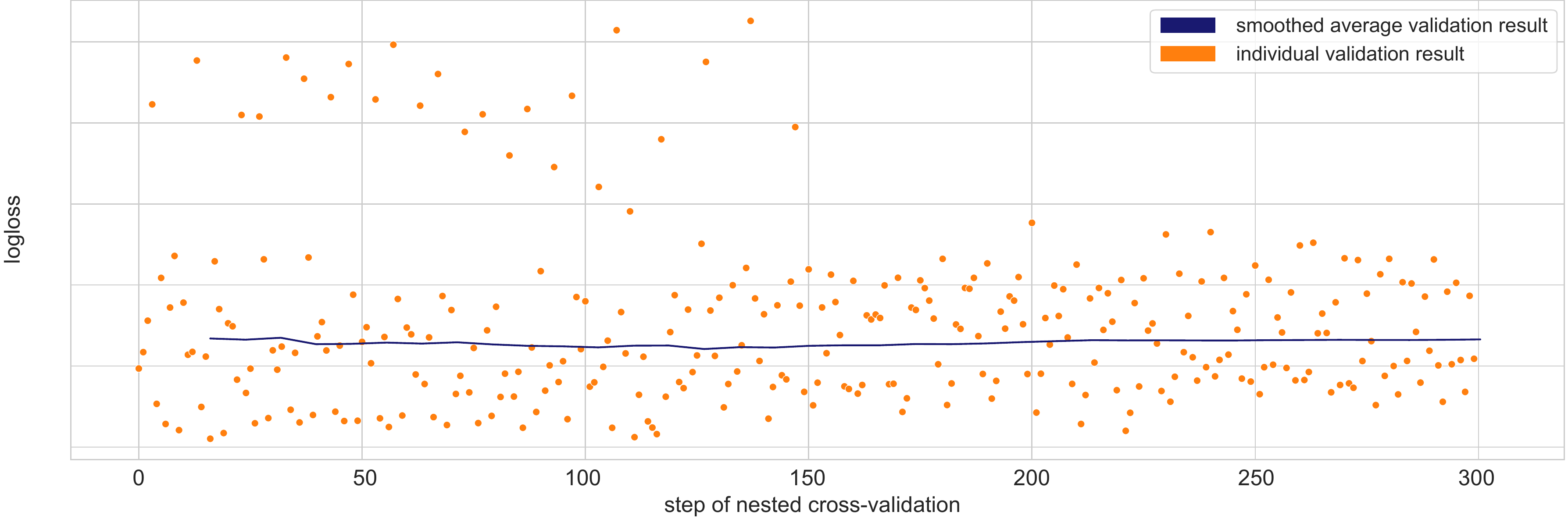}
\caption{Intermediate results of a nested cross-validation: The blue line shows the smoothed 20\% trimmed mean of the previously calculated performance evaluation metrics of a full inner cross-validation. Therefore, it starts after the first complete inner cross-validation has been fully calculated. The orange dots represent the performance evaluation metrics of every single step. This figure shows the same trial as Figure \ref{cumulated_mean}, which was also calculated with colon cancer data \cite{Alon1999} (see \nameref{experimental-setup}).} 
\label{figure_reduced_variance}
\end{figure}

The first part of the three-layer pruner is based on the comparison of intermediate results. The intermediate value in cross-validation is usually the average of the previously calculated performance measures. However, using this directly as intermediate value is unreliable for this particular use case especially at the beginning of the nested cross-validation. Due to the very small sample size and thus a tiny validation set, the inner k-fold cross-validation tends to produce a high variance in the performance evaluation metric (see Figure \ref{figure_reduced_variance}). Outliers in tiny validation sets are one reason for that. The high variance leads to an unstable mean (see Figure \ref{cumulated_mean}) and increases the probability of aborting promising trials. For this, it is necessary to adapt the calculation of intermediate values.
Hence, we do not consider each individual step of the nested cross-validation. Instead, an intermediate value always covers a complete inner k-fold cross-validation. In addition, we use the 20\% trimmed mean to smooth the high variance and eliminate the influence of outliers. These two interventions prevent overly aggressive pruning and result in more stable intermediate values (see Figure \ref{figure_reduced_variance}). Subsequently, a more stable intermediate value leads to an enhanced comparability of trials. To further increase the confidence of correct pruning, the pruner should have a high patience and prune
late. Patience in this context defines a number of iterations of the nested cross-validation
that are calculated before pruning is considered. That enables to decide based on a considerable amount of model performance evaluation data.\\
Generally, parallel processing can accelerate this computationally intensive hyperparameter optimization for high-dimensional data. In particular, asynchronous computation of the trials in a distributed environment is advantageous to reduce overhead. In this case, a pruner does not have to wait for the final results of other trials computed in parallel. Thus, multiple trials can be processed simultaneously without delay. The state-of-the-art Asynchronous Successive Halving pruner (ASHA) \cite{Li2020} applies such a strategy. It periodically monitors and ranks intermediate values and cancels the worst trials. As one component of the three-layer pruner, we have chosen a variant of ASHA implemented in Optuna \cite{Akiba2019}. This pruner scales linearly in a distributed environment. Nevertheless, other pruning strategies could also be applied that rely on the comparison of intermediate values. For example, median- or percentile-based pruning as less sophisticated pruning strategies \cite{Akiba2019}.
\subsection*{Pruning Semantically Irrelevant Trials of Hyperparameter Combinations} \label{semantic pruning}
Supplementary to the adapted calculation of intermediate values for a comparison-based pruning strategy (see \textit{\nameref{standard pruning strategy}}) the second part of the three-layer pruner considers the semantics. We distinguish between semantically irrelevant and possibly relevant trials. A trial containing a model without any selected features, as the importance of all features is zero, we define semantically irrelevant. Such a model is meaningless for the goal of feature selection. For example, a model might not include any selected features if the regularization is too strong due to inappropriate values for the hyperparameters. Trials containing those meaningless models are immediately terminated when they are detected. This pruning strategy is already active in the very first step of the nested cross-validation.
\subsection*{Pruning Trials with an Insufficient Performance Evaluation Metric} \label{threshold-pruner}
Integrating prior knowledge and accounting for variance provides an additional pruning opportunity. The third part of the tree-layer pruner does not focus on terminating inferior or semantically irrelevant trials. Instead, it prunes trials if they have an insufficient validation result compared to a user-defined threshold based on prior or domain knowledge. While comparison-based pruning requires a longer patience (see \textit{\nameref{standard pruning strategy}}), pruning against this threshold can already start earlier after half the folds of the first inner cross-validation or at least four steps. With this, we ensure to have at least four performance measurements to calculate a median. The threshold-based pruning strategy aims to close the gap at the beginning of a comparison-based standard pruning strategy (see Figure \ref{three-layer-pruner-patience}) by starting earlier. In addition, it returns intermediate results of the nested cross-validation. This has the advantage that the hyperparameter optimization algorithm can already orient itself in the hyperparameter fitness landscape without fully calculating unpromising trials.
However, we apply this pruning strategy only for the first third of all complete inner loops of a nested cross-validation. If a trial has not stopped at this point, the comparison-based pruner can make more reliable decisions.\\
As mentioned for the standard pruning strategy (see \textit{\nameref{standard pruning strategy}}), a trial should normally be pruned after considering a complete inner cross-validation to avoid overly aggressive pruning due to outliers and variance. But with this part of the three-layer pruner we already prune during the first inner k-fold cross validation if the given threshold is unlikely to be matched by its end. To achieve this, we assume that 
\begin{itemize}
\item omitting the first sample in the outer cross-validation does not massively change the median of the corresponding inner k-fold cross-validation,
\item the variance of the validation performance evaluation metrics is similar for all inner cross-validations, and
\item the first complete inner cross-validation indicates the rough level of the overall validation performance evaluation metric.\\
\end{itemize}
For this part of the combined pruning strategy, we use the median instead of the 20\% trimmed mean, as it may not be possible to omit 40\% of a very small number of results. Nevertheless, it is not useful for early pruning to directly compare the current median to the threshold, as this leads to a higher number of falsely pruned trials with respect to this threshold (see Figure \ref{percentage}). Instead, we additionally consider extrapolated values for the missing steps of the current inner cross-validation. Extrapolating all remaining missing values of the full nested cross-validation is not reasonable. This is because an overall too optimistic extrapolation could lead to a later termination of a trial, which in turn would increase the computation time.\\

For an optimistic extrapolation in order to keep promising trials, only the values that could lead to a better metric are relevant. For this reason, we only consider the deviation from the median $\tilde{x}$ in direction to optimize rather than the variance, the absolute mean or median deviation.\\
To extrapolate the missing values \textit{e}, we explored three different approaches:\\
\begin{itemize}
\item the \textbf{optimal value of the evaluation metric} (for example 0 for logloss or 1 for AUC)\\
\textit{e = optimal performance evaluation metric}\\ 
\item median $+/-$ \textbf{maximum deviation from the median} in direction to optimize\\ 
$e = \tilde{x}\begin{cases} -\text{ } max(x_1, x_2, .. , x_s) & \text{in case of minimizing}\\ +\text{ }  max(x_1, x_2, .. , x_s) & \text{in case of maximizing} \end{cases}$\\
 \\
\item median $+/-$  \textbf{mean deviation from the median} in direction to optimize\\
$e = \tilde{x}\begin{cases} -\text{ }  mean(x_1, x_2, .. , x_s) & \text{in case of minimizing}\\ +\text{ }  mean(x_1, x_2, .. , x_s) & \text{in case of maximizing} \end{cases}$\\
\\
\end{itemize}
The value, which is finally compared with the threshold, is composed of the median $\tilde{x}$ for all steps $s$ calculated so far and an optimistically extrapolated value $e$ for all steps $m$ of the current inner cross-validation that are still missing (see Equation \ref{trial-pruned}). We do not extrapolate any values for any following inner cross-validations. The number of steps already calculated $s$ is multiplied by their median $\tilde{x}$. In case of maximizing, the extrapolated value $e$ multiplied by the number of missing steps until the next complete inner cross-validation loop is added, where $m$ is the number of these steps. In case of minimization, that value is subtracted. This result divided by the number of all steps until the next complete inner cross validation  $s + m$ is compared to the threshold in order to decide if a trial is stopped.
\begin{equation}\label{trial-pruned}
\text{prune trial} = \begin{cases} threshold < \frac{(\tilde{x}*s)-(e*m)}{s + m}   & \text{in case of minimizing}\\ \\threshold > \frac{(\tilde{x}*s)+(e*m)}{s + m}  & \text{in case of maximizing} \end{cases}
\end{equation}
This part of the pruner is useful to speed up the rough search in the relevant range of the respective hyperparameters at the beginning of the optimization process. For example, Optuna \cite{Akiba2019} uses a few (10 by default) randomly chosen combinations of hyperparameter values for its start up. Therefore, those trials can often be pruned early.
\subsection*{Combined Three-Layer Pruner}\label{three-layer_complete}
\begin{figure}
\begin{center}
\includegraphics[width=122mm]{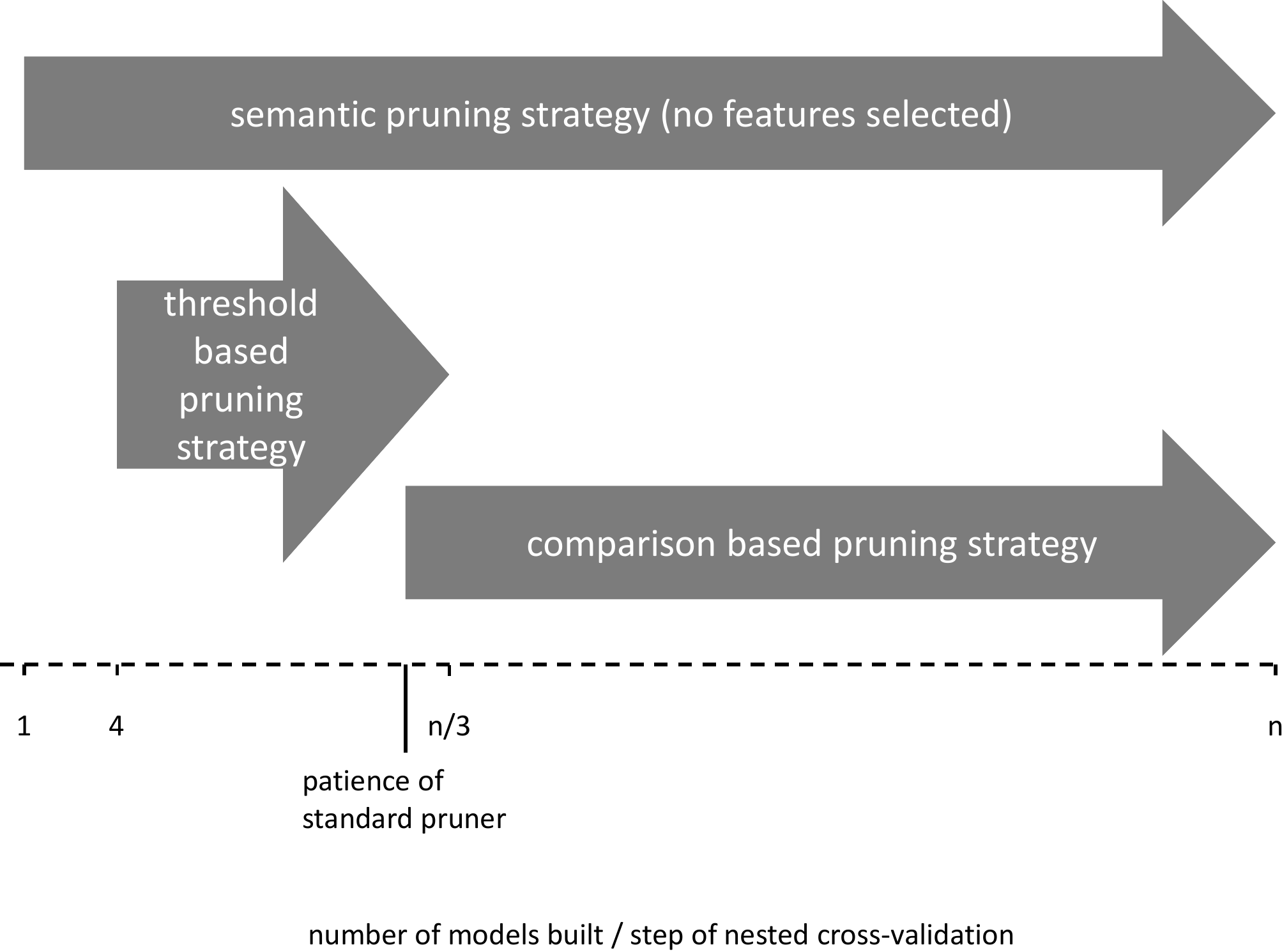}
\caption{Operating times of the different parts of the three-layer pruner.}
\label{three-layer-pruner-patience}
\end{center} 
\end{figure}

The elements of the combined three-layer pruner operate within different ranges of the nested cross-validation (see Figure \ref{three-layer-pruner-patience}). Applying the semantic pruning strategy, a trial is pruned immediately if no features are selected, starting with the very first model created. In contrast, the threshold-based pruning strategy becomes active only after at least 4 steps or half of the folds of the first inner cross-validation loop. It is only active within the first three iterations of the outer cross-validation loop. After that, the more reliable standard comparison-based pruning strategy takes over. It removes all trials worse than previous ones or less promising than those currently running in parallel. This comparison-based strategy requires computing significantly more steps if there is a high variance in individual results in order to allow a valid comparison of different trials. A decision based on more validation results reduces the risk of stopping a promising trial.
The two pruning strategies, based on semantic and prior knowledge, respectively, complement this standard pruning strategy to further speed up hyperparameter optimization (see Figure \ref{percentage}).\\
When a comparison-based or an extrapolating threshold based pruning strategy terminates a trial, the current evaluation metric should be returned. This has the advantage that the hyperparameter optimization algorithm receives a representative signal in order to be able to adapt better.

\section*{Experimental Setup} \label{experimental-setup}
We repeated and evaluated three experiments to compare the pruning strategies. They are based on three different biological data sets. Whereas other high-dimensional biological datasets with a small sample size most likely yield similar results.
%We expect other high-dimensional biological datasets with a small sample size to provide similar results. 
Each individual experiment includes 30 repetitions of the same hyperparameter optimization with 40 trials each. All trials are fully computed to allow simulations and direct comparisons.\\
In contrast, the fourth experiment implements the principle of returning the current performance evaluation metric when stopping. It not only simulates, but shows the interaction of the three-layer pruner as a whole. We use only one biological data set and one hyperparameter optimization for this experiment since other examples follow a similar pattern.\\
All computations were performed on a cluster platform (Sun Grid Engine) using twenty cores working in parallel. Results and intermediate results were saved in a local SQLite database with exclusive access for the current experiment.
\subsection*{Data Sets used for Empirical Evaluation} \label{data}
 
%"The dataset(s) supporting the conclusions of this article is(are) included within the article (and its additional file(s))." 
% 
The first biological data set from Alon et al. includes 40 tumor and 22 normal samples from colon-cancer patients with 2000 measured genes \cite{Alon1999}. The second data set is based on a prostate cancer study comparing 52 patients with 50 healthy controls. The genetic activity was measured for 6033 genes \cite{Efron2016}. Golub et al. provide the third data set based on gene expression measurements on 72 leukemia patients with 7128 genes \cite{Golub1999}. 47 patients with acute lymphoblastic leukemia (ALL) and 25 patients with acute myeloid leukemia (AML) \cite{Camargo2020}.\\ 
We transformed all data sets so that each row corresponds to a patient, each column to a gene, and the label as an integer 0 or 1 to the respective classes. In addition, to ensure comparability and to simulate pilot studies with very small sample sizes, we included only the first 15 samples of each class. Thus, all data sets are balanced and contain 30 samples. Since there are no missing values and scaling is not required for tree-based models, no further preprocessing was performed.
\subsection*{Scoring Metric}
For our experiments with balanced binary classes, we decided to use logloss as a performance evaluation metric.  We chose this way because metrics based on the confusion matrix \cite{Tharwat2021}, such as F1, accuracy, average precision value, and even AUC, have the disadvantage that the results are not continuous when minimal validation sets are used. Logloss, on the other hand, provides a more fine-grained signal \cite{Bishop2006} even for these tiny validation sets.
\subsection*{Hyperparameter Optimization Setup}\label{hyperparameter-optimization}
We used Optuna v2.9.1 \cite{Akiba2019} as automatic hyperparameter optimization software for our experimental setup. It is easy to use, scales well in distributed systems, integrates with Sun Grid Engine, and already provides an implementation of the Asynchronous Successive Halving (ASHA) pruner \cite{Li2020}. We opt for the multivariate  Tree-structured Parzen Estimator (TPE) sampler, which outperforms the independent TPE sampler, as Falkner et al. have shown \cite{Falkner2018}.\\
We evaluate each combination of hyperparameter values with nested cross-validation to obtain an unbiased performance evaluation. For the outer loop, we use leave-one-out cross-validation to save as much data as possible for the training set.
As inner loop, we use a stratified 10-fold cross-validation\footnote{The calculation time would further increase if the even more suitable leave-pair-out cross-validation \cite{Airola2009} was used instead of the standard 10-fold inner cross-validation. If regression was chosen instead of classification, nested leave one out cross-validation would also be possible to save even more data for the validation set.} 
including at least one sample of each class in each validation set.  
All intermediate validation results of each fold are stored for later analysis. While we did not calculate the test metric for our pruning simulations, obviously in practice this is mandatory. To predict the associated class of an outer folds test set, each model of an inner fold should be used.\\ 
We use random forest for our simulations of tree-based embedded feature selection \cite{Qi2012}. Tree-based supervised learning performs well with unstandardized raw data and is robust to outliers. The models for our simulation we have trained with the fast C++ random forest implementation of lightGBM v3.2.1 \cite{Ke2017}. We set a fixed number of 100 trees to enhance comparability of trial durations. As further hyperparameters for regularization we choose \verb|lambda_l1|, \verb|min_gain_to_split| and \verb|min_data_in_leave| to combat overfitting. The complete hyperparameter space and the parameters for the asynchronous successive halfing pruner are listed in \nameref{app:A}. For semantic pruning (see \textit{\nameref{semantic pruning}}), we determine the feature importances based on information gain \cite{Quinlan1986}, the total gain of the splits that use the feature \cite{lightgbmBoosterDoc}.
\subsection*{Comparison of Pruning Strategies}
Due to the random initialization of hyperparameter optimization and training, it is not realistic to take accurate measurements for a direct comparison of the pruners. Therefore, on the one hand, we repeat each hyperparameter optimization thirty times. On the other hand, we simulate optimizations based on previously stored performance evaluation metrics. To compare the different pruning strategies with consistent data, all respective trials were calculated to completion. The principle of returning the current performance evaluation metric instead of terminating the experiment was not implemented in these experiments. In this way, we exclude the random part of the model building process. This allows to directly compare different extrapolating threshold pruning strategies on the same basis.\\
Pruning deals with a trade-off between the early termination of trials and the risk of pruning a promising trial. Accordingly, we compare pruning strategies regarding 
\begin{itemize}
\item[a)] the number of steps until pruning,
\item[b)] the number of incorrectly stopped trials, and
\item[c)] their margin of error.
\end{itemize}
With the margin of error we evaluate the falsely pruned trials not only quantitatively but also qualitatively.
For the direct comparison, however, we did not consider time measurements because the experiments ran on different hardware, which leads to a lack of comparability. However, the number of models built is proportional to the calculation time and allows an objective comparison in this context. Therefore, we assume that the calculation of fewer steps within a nested cross-validation requires proportionally less time.

\section*{Results}
\begin{figure}[h!]
\begin{center}
\includegraphics[width=122mm]{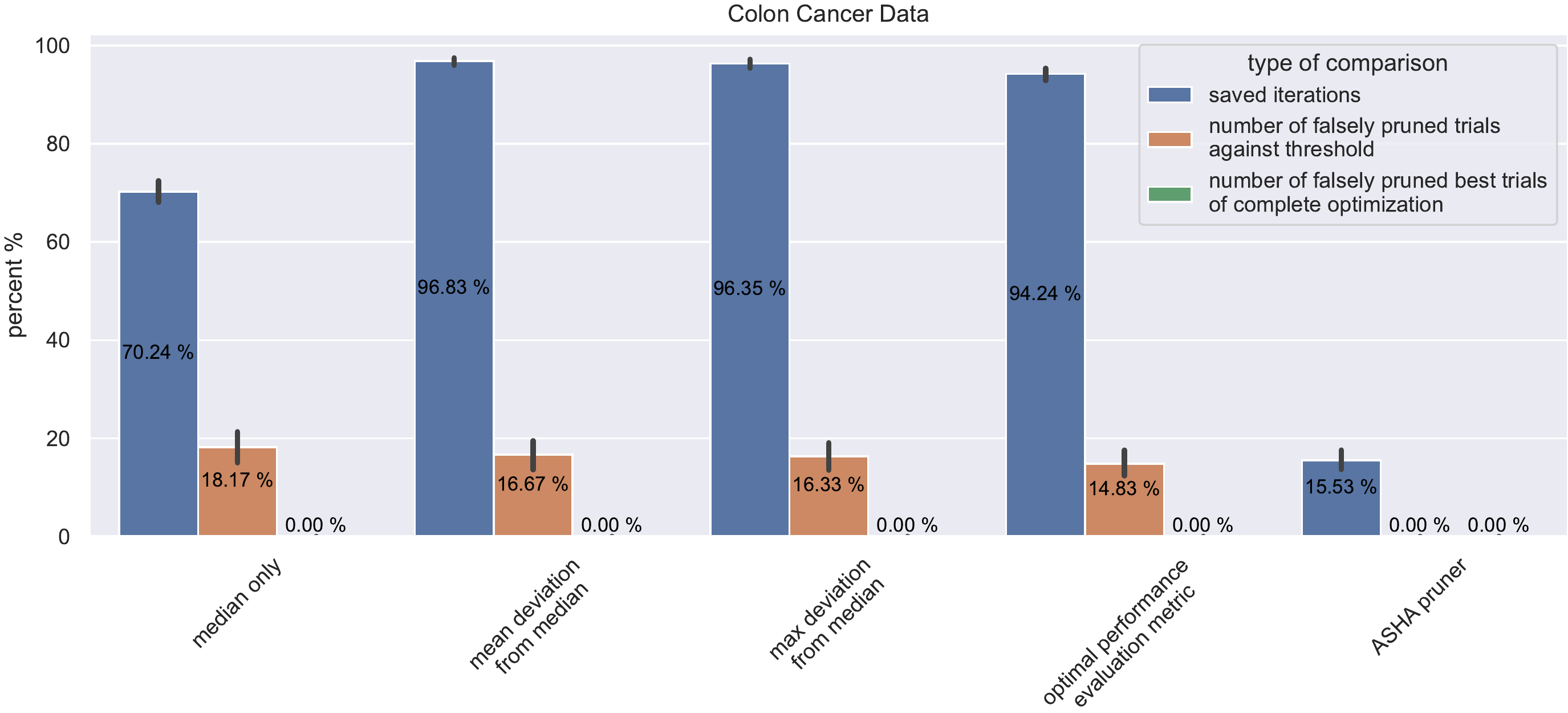}
\includegraphics[width=122mm]{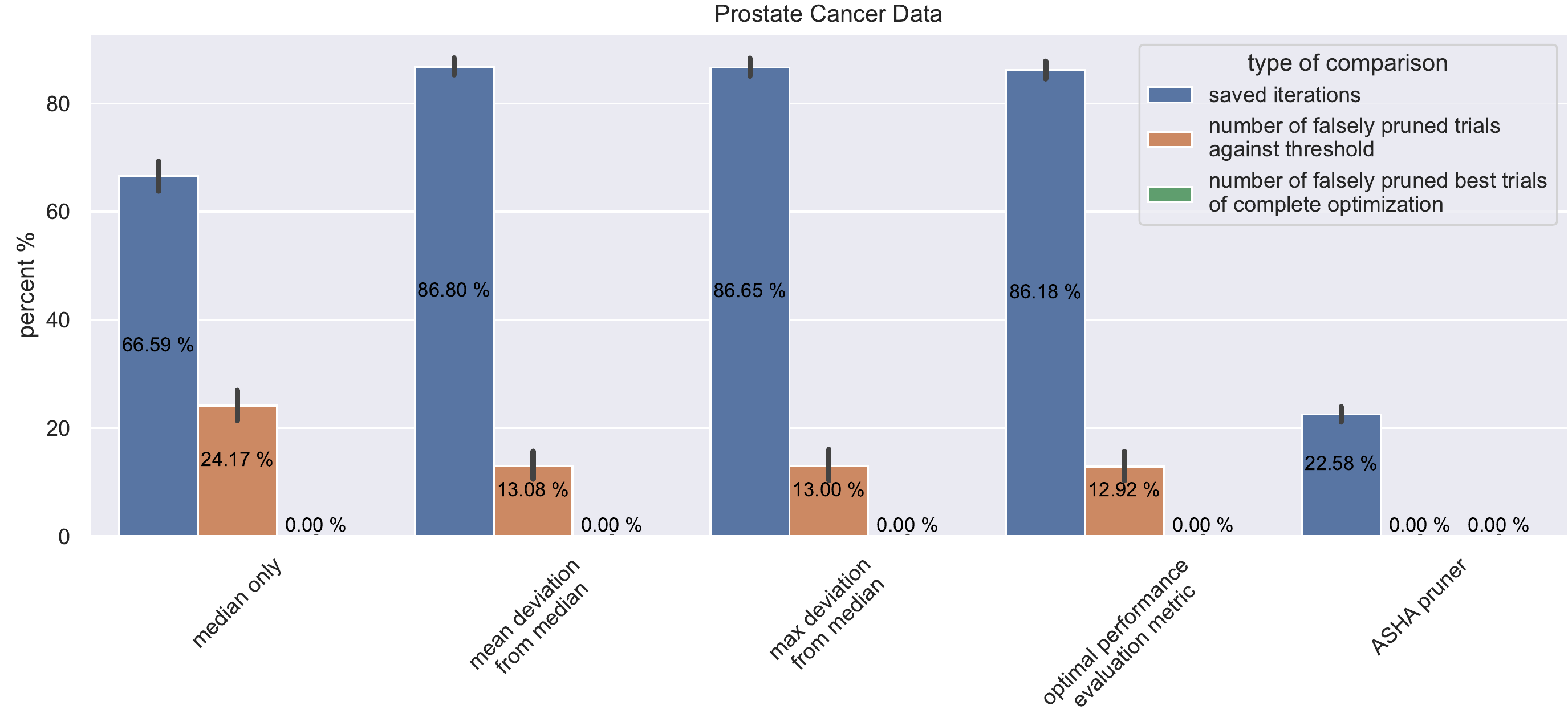}
\includegraphics[width=122mm]{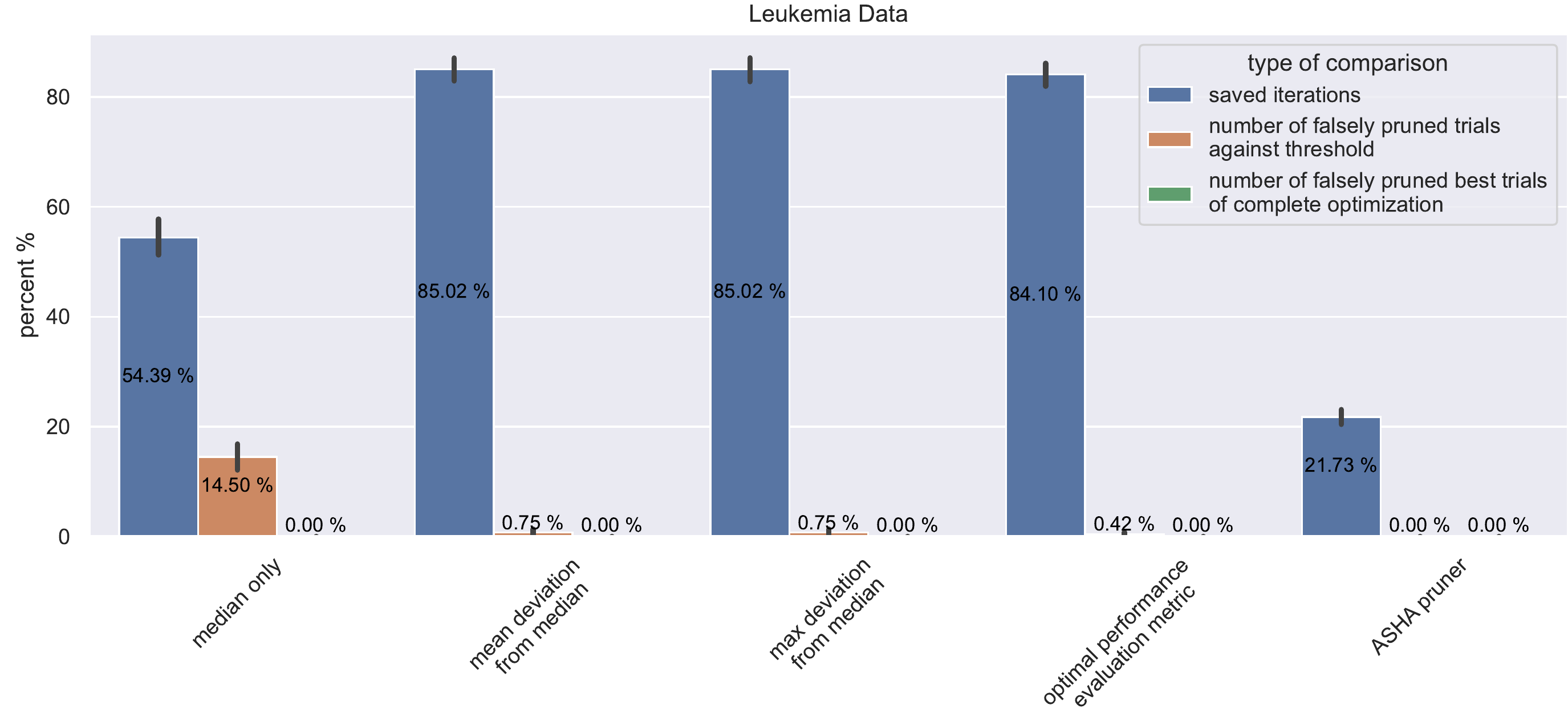}
\caption{The different extrapolation strategies within the three-layer pruner compared to using the ASHA pruner alone as a single pruner component applied to three different biological data sets. The error bars show the deviation between the repetitions of each experiment (see \textit{\nameref{experimental-setup}} for further details)}.
\label{percentage}
\end{center}
\end{figure}

\begin{figure}[h!]
\includegraphics[width=106mm]{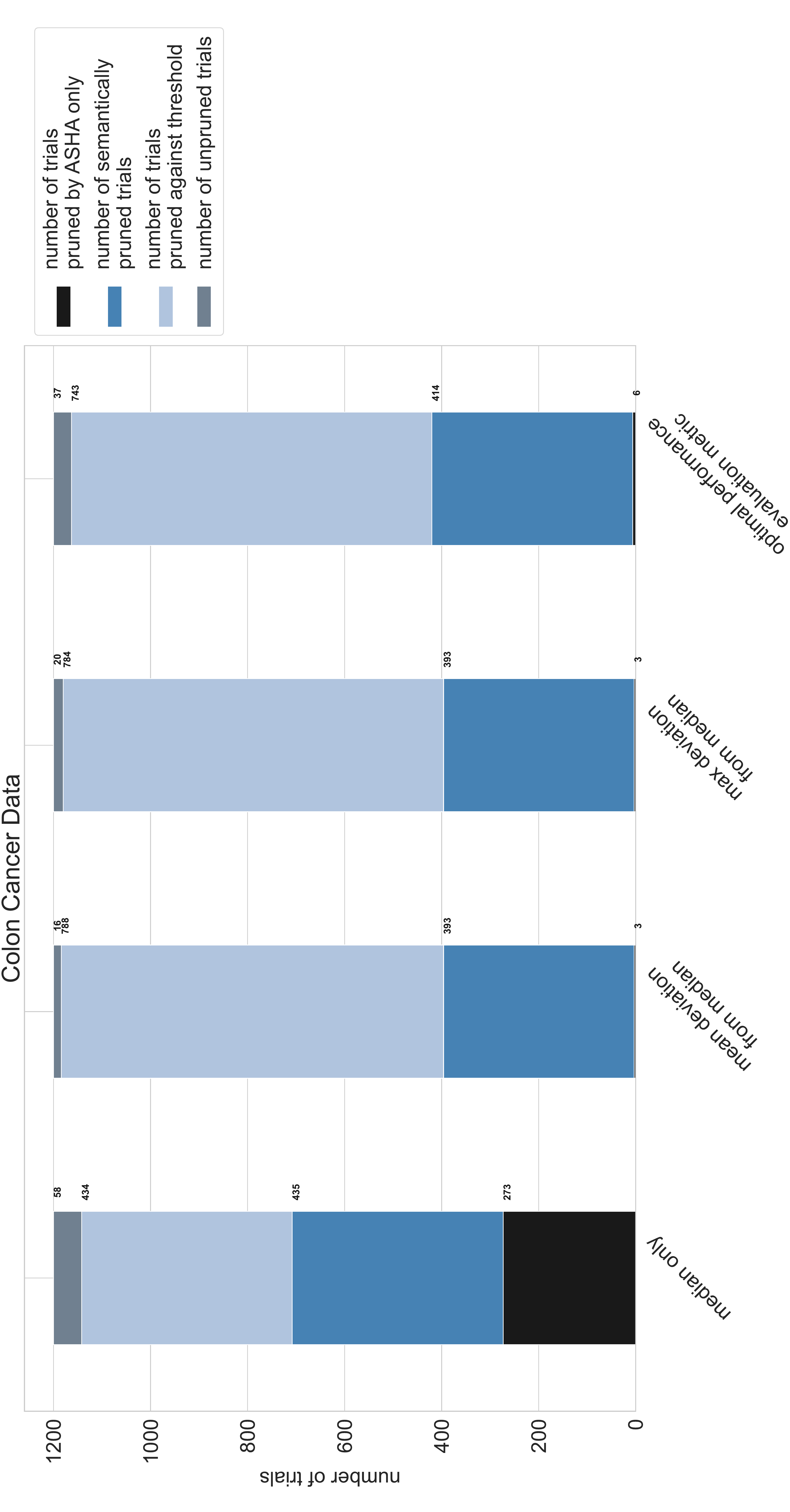}
\caption{\textbf{Comparison of different extrapolation strategies}: Combined parts of the three-layer-pruner for 30 repetitions of hyperparameter optimization for colon cancer data analysis containing 40 trials each.} 
\label{combination1}
\end{figure}

\begin{figure}[h!]
\includegraphics[width=106mm]{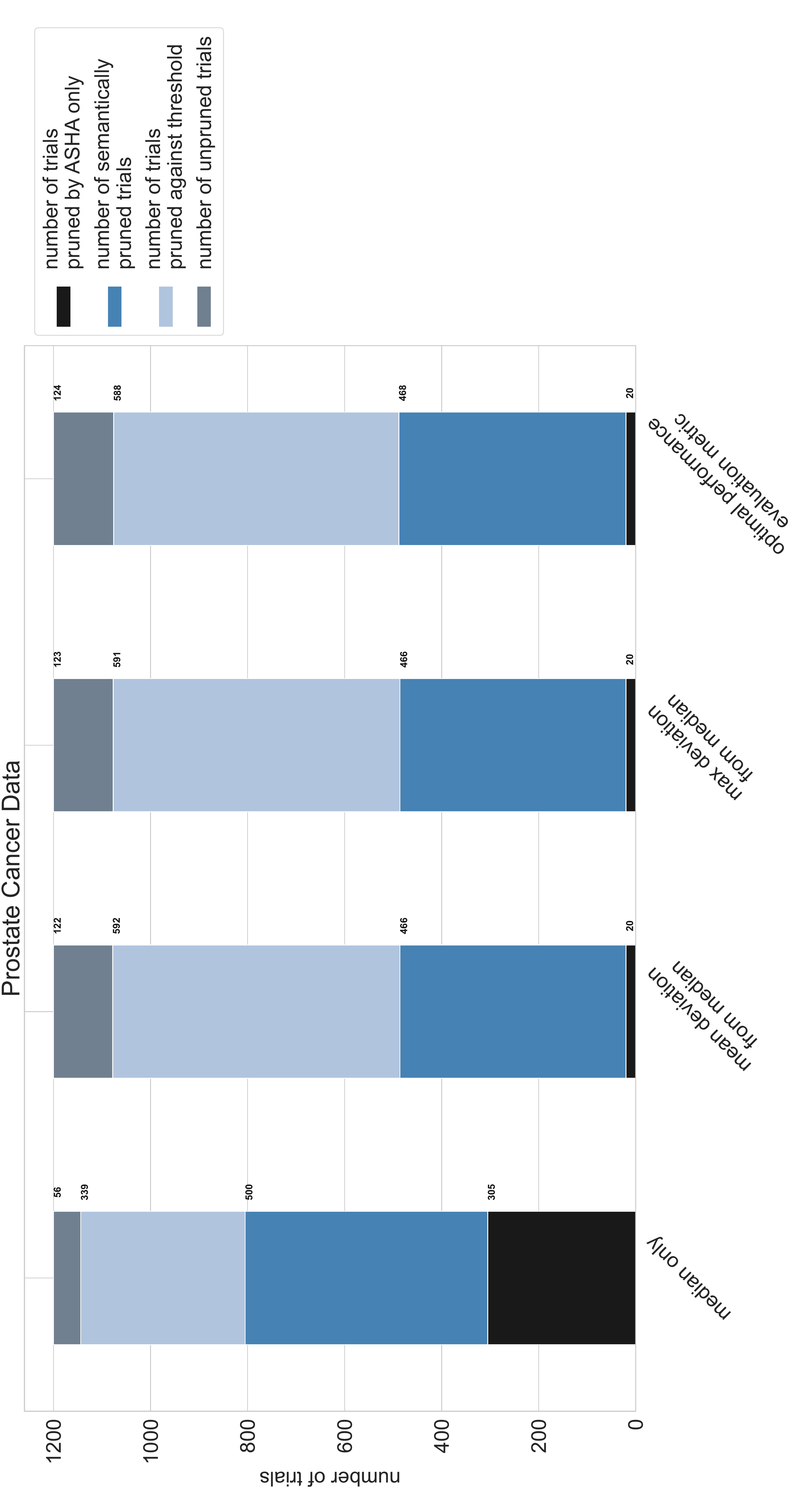}
\caption{\textbf{Comparison of different extrapolation strategies}: Combined parts of the three-layer-pruner for 30 repetitions of hyperparameter optimization for prostate cancer data analysis containing 40 trials each.} 
\label{combination2}
\end{figure}

\begin{figure}[h!]
\includegraphics[width=106mm]{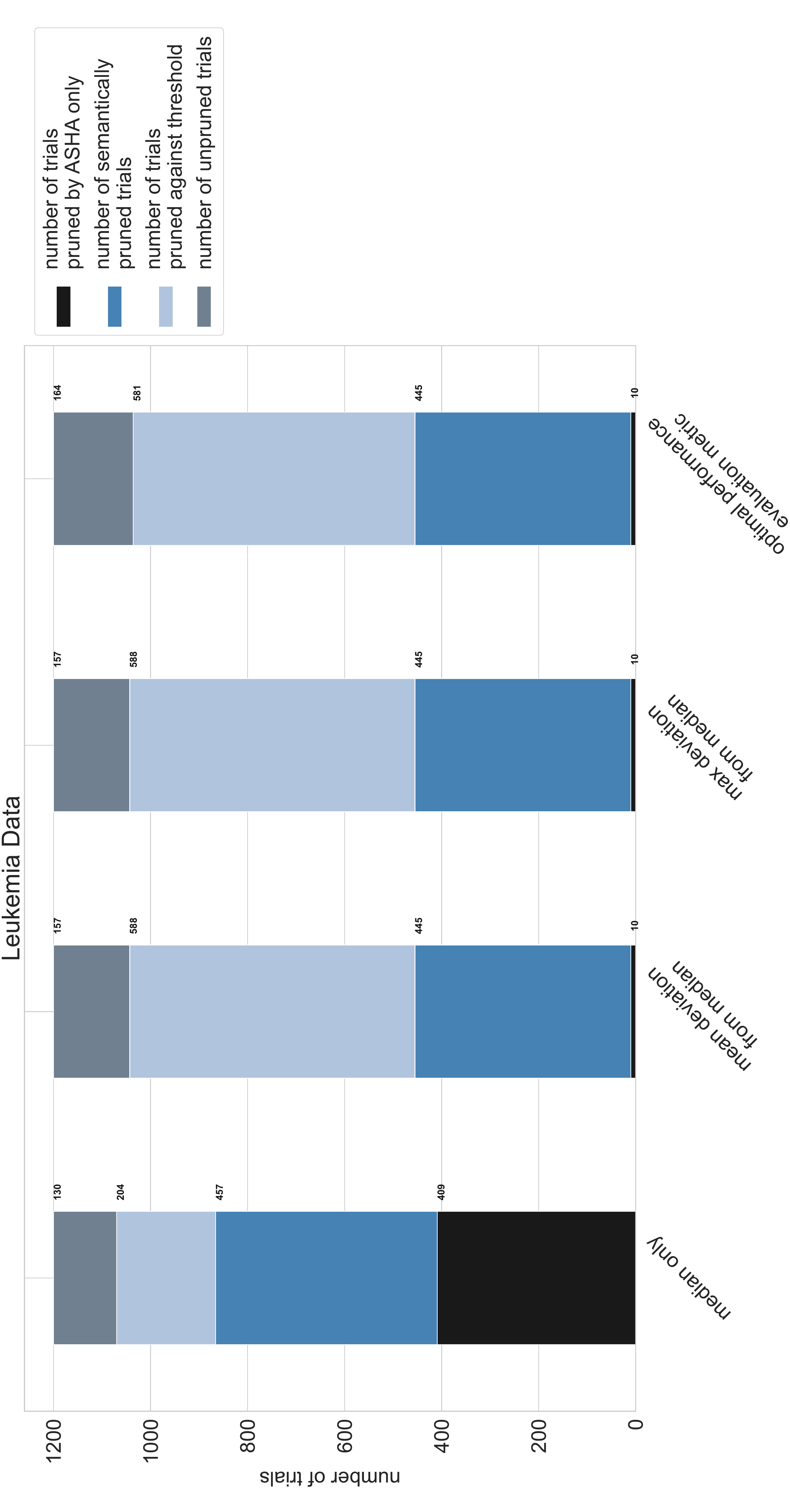}
\caption{\textbf{Comparison of different extrapolation strategies}: Combined parts of the three-layer-pruner for 30 repetitions of hyperparameter optimization for leukemia data analysis containing 40 trials each.} 
\label{combination3}
\end{figure}

\begin{figure}[h!]
\begin{center}
\includegraphics[width=122mm]{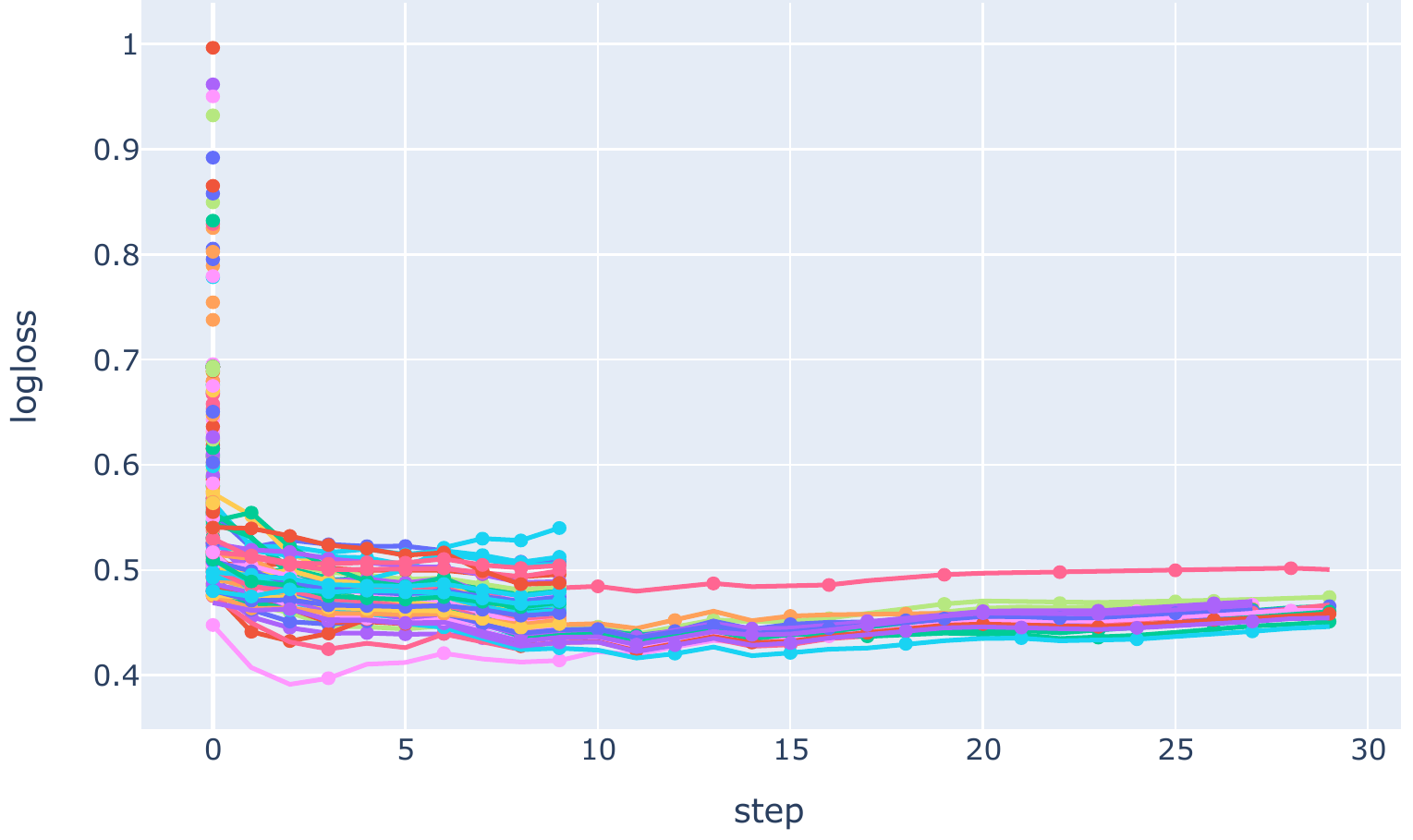}
\caption{Pruning unpromising trials with the combined three-layer pruner: Terminated trials based on the intermediate values of a hyperparameter optimization (colon cancer data \cite{Alon1999}). The steps represent the iterations of the outer cross-validation.}
\label{three-layer-pruner}
\end{center}
\end{figure} 

\begin{figure}[h!]
\includegraphics[width=122mm]{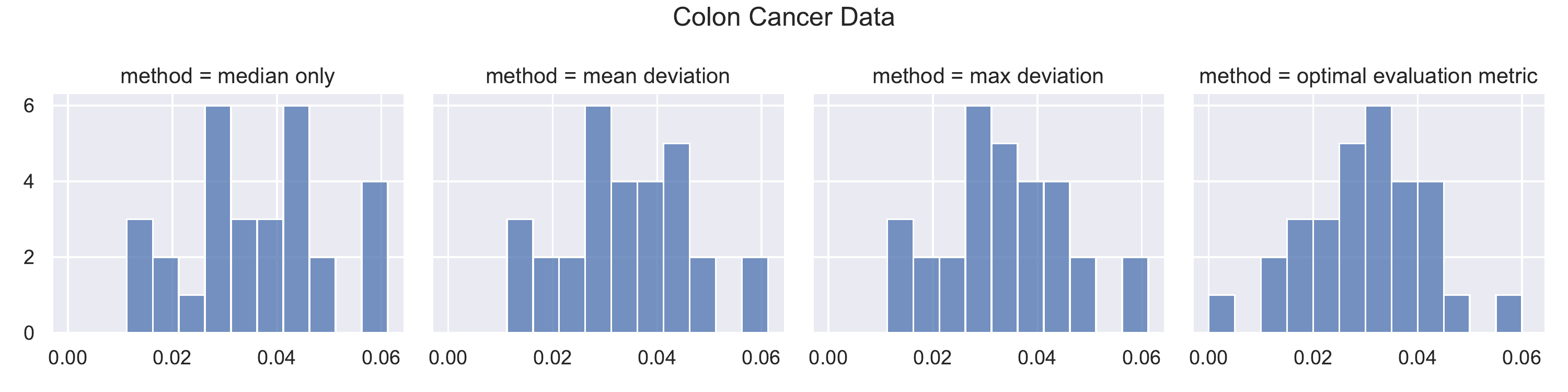}
\\

\includegraphics[width=122mm]{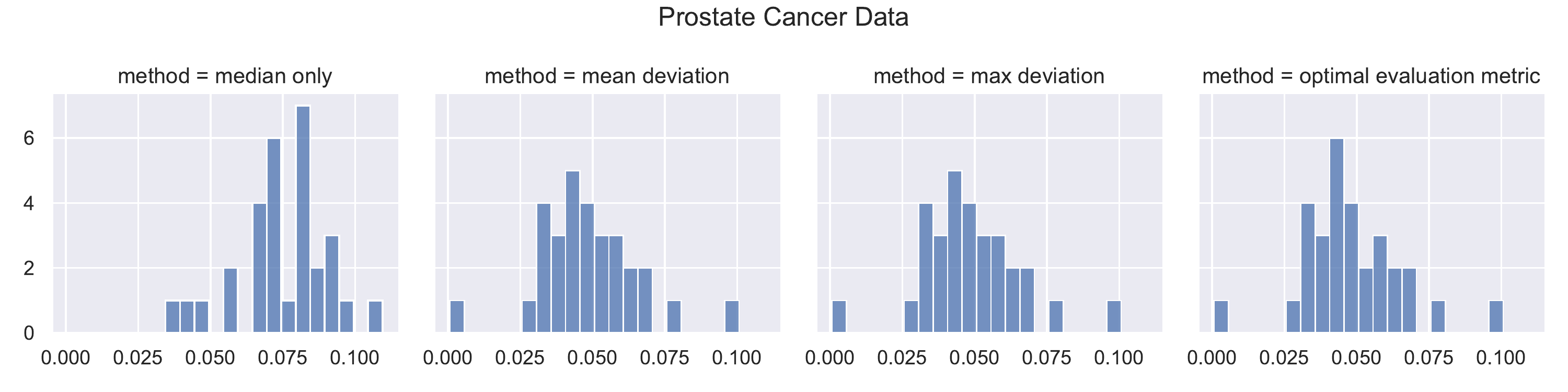}
\\

\includegraphics[width=122mm]{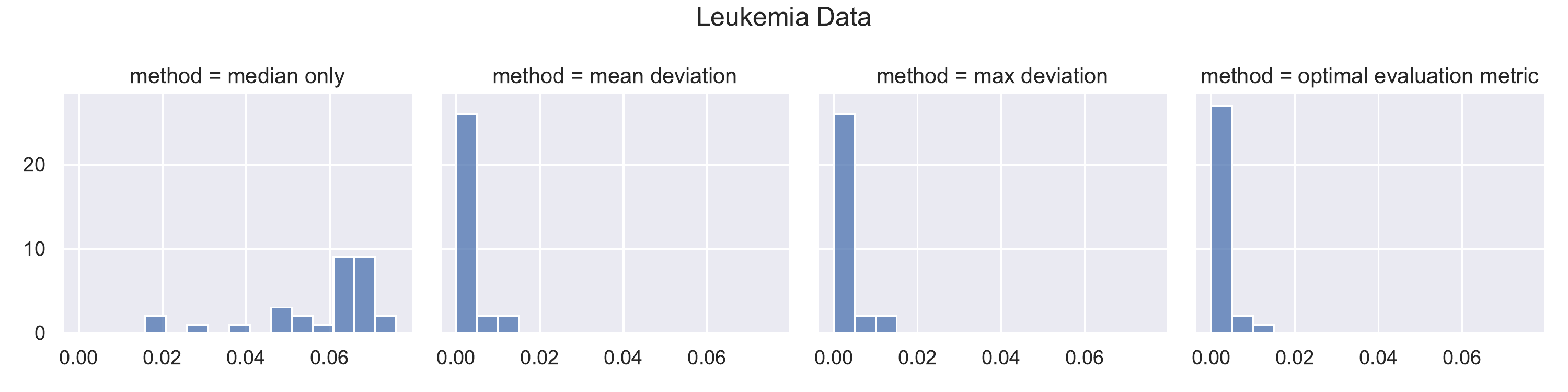}
\caption{\textbf{Margin of error for incorrectly pruned trials compared to the set threshold}: On the x-axis the difference between the 20\% trimmed mean of the performance evaluation metrics (logloss) of a full nested cross-validation and the threshold for the extrapolating pruner is shown. Only falsely pruned trials are included. Their respective number is plotted on the y-axes. For one hyperparameter optimization, only the globally largest margin of error was considered. Four different extrapolation methods are compared: 
1. no extrapolation only the current median 2. mean deviation from the median subtracted from the current median 3. maximum deviation from the median subtracted from the current median 4. the optimal performance evaluation metric. 
The figures apply to a complete experiment with 30 repeated hyperparameter optimizations including 40 trials each. Three biological data sets were examined using the same experimental setup. Further experimental details are specified in \textit{\nameref{experimental-setup}}.}
\label{qualitative_comparison}
\end{figure}
In our repeated experiments, the best overall trial was never pruned. At the same time, considerably fewer models had to be trained (see Figure \ref{percentage}) proportionally saving a remarkable amount of time.
%while saving a remarkable amount of time.
All three parts of the combined three-layer pruner contributed to this acceleration (see Figures \ref{combination1},\ref{combination2} and \ref{combination3}). Not only the adapted usage of the state-of-the-art comparison-based ASHA pruner \cite{Akiba2019} was responsible for this (see \textit{\nameref{standard pruning strategy}}).
We can show that the additional pruning strategies based on semantic or prior knowledge can prevent computing complete trials or larger parts of trials that might not lead to a useful result. Figure \ref{three-layer-pruner} illustrates the early termination of unpromising trials within the first inner cross-validation combined with the comparison-based pruning of acceptable trials.\\
All three extrapolation methods for the threshold pruning strategy accelerate and enhance the hyperparameter optimization. They result in more correctly completed trials and fewer trained models than comparing a threshold directly to the median (see Figure \ref{percentage}). Extrapolation compensates the uncertainty due to the lack of robustness of the results at the beginning of the nested cross-validation. The \textit{optimal value of the evaluation metric} as a basis for extrapolation leads to the smallest number of falsely pruned trials and to the highest number of models to be trained. On the other hand, the \textit{maximum deviation from the median} in direction to optimize leads to a slight increase in the number of incorrectly pruned trials but, in turn, reduces the calculation time. The \textit{mean deviation from the median} in direction to optimize finally requires the least number of trained models for the hyperparameter optimization but has another tiny increase in the number of falsely pruned trials. 
\textit{Mean deviation from median} as extrapolation method for the threshold pruning strategy was the approach with the highest acceleration. 
Using this extrapolation method, the combined three-layer pruner could still achieve the same optimization result as the standard ASHA pruner \cite{Akiba2019} as single pruning strategy. But on average, 81.3\% fewer models for the colon cancer data \cite{Alon1999}, 64.2\% fewer models for the prostate cancer data \cite{Efron2016} and 63.3\% fewer models for the leukemia data \cite{Golub1999} had to be trained. The corresponding error bars in Figure \ref{percentage} show that the deviations of the acceleration between the repetitions of the individual experiments are relatively small. This indicates a repeatable and stable acceleration even though the hyperparameter optimization is initialized randomly.\\
Furthermore, the three extrapolation strategies and the threshold pruner without extrapolation were qualitatively compared. This qualitative analysis of the incorrectly pruned trials for three different biological datasets is shown in Figure \ref{qualitative_comparison}. It is again noticeable that larger errors occur when only the current median without any extrapolation is compared against a threshold. As expected, the correctness is the highest for the optimal performance evaluation metric as basis for the extrapolated values. \\
Despite any (expected) errors, a threshold with a clear distance to a likely best performance evaluation outcome always led to the completion of the most promising trial of the hyperparameter optimization. Due to this distance, small errors related to the threshold were not relevant for the global outcome of the complete hyperparameter optimization.
\section*{Discussion}
The overall goal of the combined pruning strategy is calculating many parallel trials, stop unpromising ones early, and continue the best. Termination of unpromising trials based on the principle of extrapolation and semantic pruning can also accelerate simple cross-validation. However, speedup might be smaller in this case, since normally the overall number of iterations in cross-validation is lower compared to nested cross-validation. For nested cross-validation these two pruning strategies based on semantic or prior knowledge help to roughly tune the hyperparameters fast. They increase the speed-up compared to using a comparison-based pruning strategy alone.  Especially at the beginning of the hyperparameter optimization, comparison-based standard pruners compute complete trials or larger parts of trials that do not lead to a useful result. Otherwise, they  would not have a stable baseline for comparison, which in turn would risk stopping promising trials. The advantage of the comparison-based pruning strategy is the ability to finely compare superior trials. Comparison-based pruning is, of course, not limited to tree-based methods. It can also accelerate the hyperparameter optimization of any embedded feature selection or supervised learning method that provides intermediate results, such as LASSO regression \cite{Muthukrishnan2016}.\\
%By using extrapolation when pruning against a threshold, more trials are completed correctly and fewer models need to be trained than when only comparing the current median or the even more unstable mean. This is because the extrapolation compensates the uncertainty caused by the lack of robustness of the performance evaluation results at the beginning of the nested cross-validation. 
However, saving time by early termination of trials must always be balanced against the risk of pruning a promising trial. For this reason, the pruning strategy based on a user-defined threshold takes prior knowledge into account. The threshold itself can influence the global error probability, but finding the best threshold is a challenge. The closer it is to the optimum, the more aggressive the pruning strategy and the higher the probability of error, and vice versa. The  threshold should be an unacceptable value or a minimum requirement for the evaluation metric in the specific domain. Therefore, it optimally should be chosen with human experience. Likewise, knowledge from previous analyses related to the expected evaluation metric can serve as a base value. If determining a threshold in this way is not possible, it could be set slightly better than the naive classification baseline for the given data. Another option to get orientation regarding the expected evaluation metric could be applying an appropriate filter method for feature selection \cite{Bommert2020}. However, a single good biomarker candidate or feature might appear only by chance in high-dimensional data with tiny sample size. To better estimate this probability, the HAUCA method \cite{Klawonn2016} could be applied. It provides information on whether more informative features are likely to be present in the data than would be expected by chance for a given value of an evaluation metric.\\
Another conceivable option would be to automatically adjust the threshold to the current best performance evaluation metric after each successful trial. In practice, however, this can lead to pruning of promising trials, as it is likely that the threshold will be slightly mismatched (see Figure \ref{qualitative_comparison}). These mismatches can occur due to a high variance of the performance evaluation metric. We therefore suggest choosing a defensive threshold that has a sufficient distance to the expected best value of the performance evaluation metric.\\
In addition to the threshold, the method of extrapolation affects the acceleration as well as the probability of error. The more "aggressive" (e.g. \textit{mean deviation from the median}) the method of extrapolation, the less models are built but the higher the risk of falsely pruned trials. In contrast, the more "defensive" (e.g. \textit{optimal value of the evaluation metric}) the method of extrapolation, the more models must be calculated and the lower the risk of error.\\ 

Additional speed-up can be achieved by parallelization. As the combined asynchronous pruning strategy scales very well in distributed systems, massive parallelization of trials can further reduce the computation time. 
%This is most effective when training is computationally expensive. However, when resources are limited, parallel execution of trials is usually more efficient than parallel training. The reason for this is that the long computation time for nested cross-validation results mainly from the relatively large number of models to be trained and not from the long training times of the individual models.
\section*{Conclusion and Outlook}
Hyperparameter optimization with high-dimensional data and nested cross\hyp{}validation leads to long computation times. Thus, its speed-up is a key factor. Combining three different specifically adapted pruning strategies enables a substantially acceleration of the hyperparameter optimization. The part of the combined pruning strategy based on a user-defined threshold incorporates prior knowledge, which might be challenging to determine.\\
A complementary, time-saving early stopping approach is proposed by Makarova et al. \cite{Makarova} which focuses on early stopping of the complete hyperparameter optimization to avoid overfitting due to too extensive hyperparameter optimization. Combining both approaches is a promising future research topic.\\
In bioinformatics, for example, only accelerated hyperparameter optimization enables the analysis of larger multi-omics or very large high-throughput datasets in a reasonable time. In particular, faster computation allows the application of advanced ensemble techniques - the combination of different feature selection approaches. These ensemble techniques, in turn, can provide more reliable results \cite{Luftinger2021}. From the infection research perspective, the improved reliability increases the likelihood of finding more relevant biomarker candidates. And this allows cutting-edge research to gain new insights.\\
Reduced computation time can not only enable or speed up research in general, but also reduce costs such as personnel, hardware, management, cloud service, or energy costs. To slow down global warming, it is desirable to contribute to the reduction of CO$_2$ emissions by improving algorithms.\\

\section*{Availability and Requirements}\label{sec:Requirements}

     Project name: cv-pruner\\
     Project home page: https://github.com/sigrun-may/cv-pruner\\
     Operating system(s): Platform independent\\
     Programming language: Python\\
     Other requirements: Python 3.8 or higher\\
     License: MIT\\
     Any restrictions to use by non-academics: -
     
\section*{Appendix A.}\label{app:A}
In this appendix, first we describe the tuned hyperparameters for regulation to combat overfitting. Following this, we present the complete hyperparameter space for the experiments (Table \ref{tab:hyperparameter-space}) and the paramenters for the asynchronous successive halfing pruner (Table \ref{tab:hyperparameters-asha}) from Chapter \textit{\nameref{hyperparameter-optimization}}.\\
\noindent In addition to the parameters required for lightGBM, we tuned the following supplemental hyperparameters for regulation to prevent overfitting:
\begin{itemize}
\item \verb|lambda_l1|: L1 regularization.
\item \verb|min_gain_to_split|: Gain is basically the reduction in training loss that results from adding a split point.
\item \verb|min_data_in_leave|: Minimum number of observations that must fall into a tree node for it to be added.
\end{itemize}
Further details can be found at \url{https://github.com/microsoft/LightGBM}\\ 

\begin{table}[h!]
\caption{Hyperparameters of experiments taken from uniform distribution.}
\label{tab:hyperparameter-space}
  \begin{tabular}{lcc} 
    \hline
    hyperparameter & min value & max value \\  \hline 
	min data in leaf & 2 & $\left \lceil{sample \ size / 2}\right \rceil$ \\ 
	lambda l1 & 0.0 & 3.0 \\ 
	min gain to split & 0 & 5 \\ 
	max depth & 2 & 15 \\ 
	bagging fraction & 0.1 & 1.0 \\ 
	bagging freq & 1 & 10 \\
	num leaves & 2 & $\min(2^{max \ depth} - 1, \quad 80)$ \\ \hline
  \end{tabular}
\end{table}

\begin{table}[h!]
\caption{Parameters for the asynchronous successive halving pruner.}
\label{tab:hyperparameters-asha}
  \begin{tabular}{lcc} 
      \hline
    hyperparameter & value \\   \hline 
	min resource & auto \\ 
	reduction factor & 3  \\ 
	min early stopping rate & 2  \\ 
	bootstrap count & 0 \\ \hline
  \end{tabular}
\end{table}

%%%%%%%%%%%%%%%%%%%%%%%%%%%%%%%%%%%%%%%%%%%%%%
%%                                          %%
%% Backmatter begins here                   %%
%%                                          %%
%%%%%%%%%%%%%%%%%%%%%%%%%%%%%%%%%%%%%%%%%%%%%%
\begin{backmatter}
\section*{Acknowledgements}%% if any
Diana Zimper, Philip May, Juliane Hoffmann

\section*{Funding}%% if any
We would like to acknowledge funding for this project
from "Paving the way towards individualized vaccination (i.Vacc) - Exploring multi-omics Big Data in the general population based on a digital mHealth cohort". The funding agency did not influence the design of the experiments, analysis and interpretation of the data, or writing of the manuscript.

\section*{Abbreviations}%% if any
ASHA - Asynchronous Successive Halving: "ASHA is an asynchronous pruning strategy suitable for massive parallelism \cite{Li2020}. Successive halving is a bandit-based algorithm to determine the best among multiple configurations \cite{Akiba2019}.
%"Successive Halving is a bandit-based algorithm to identify the best one among multiple configurations." \cite{} \cite{Akiba2019}
%%https://optuna.readthedocs.io/en/stable/reference/generated/optuna.pruners.SuccessiveHalvingPruner.html#optuna.pruners.SuccessiveHalvingPruner 
%ASHA is a robust successive halving algorithm suitable for massive parallelism. \cite{Li2020} 

\section*{Availability of data and materials}%% if any
All data analysed are included in this published articles \cite{Alon1999}, \cite{Efron2016}, \cite{Golub1999}.

\section*{Ethics approval and consent to participate}%% if any
Not applicable.

\section*{Competing interests}
The authors declare that they have no competing interests.

\section*{Consent for publication}%% if any
Not applicable.

\section*{Authors' contributions}
S.M. designed the algorithm, analyzed and interpreted the data and created new Software used in the work. S.H. and F.K. substantively revised the manuscript. All authors read and approved the final manuscript.

%%  After submission of the .TEX file,                     %%
%%  you will be prompted to submit your .BBL file.         %%

% if your bibliography is in bibtex format, use those commands:
\bibliographystyle{bmc-mathphys} % Style BST file (bmc-mathphys, vancouver, spbasic).
\bibliography{pruner-bmc}      % Bibliography file (usually '*.bib' )

\end{backmatter}
\end{document}